\documentclass[letterpaper,journal]{IEEEtran}
\usepackage{amsmath,amsfonts}
\usepackage{array}
\usepackage{textcomp}
\usepackage{url}
\usepackage{verbatim}
\usepackage{graphicx}
\usepackage{cite}
\usepackage{placeins}
\usepackage{float}
\begin{document}

\title{MVG-KAN: Multi-View Geo-Wind Guided KAN for PM$_{2.5}$ Forecasting}

\author{Cheng Huang, Muyao Guan, Jairus Yougui Railey, Ning Xu, Honghui Xu, Changjiang Zhang, Zhen Zhang, Shiqing Zhang, and Cong Bai%
	\thanks{Cheng Huang, Muyao Guan, Jairus Yougui Railey, Ning Xu, Honghui Xu, Changjiang Zhang, Zhen Zhang, and Shiqing Zhang are with the College of Artificial Intelligence, Taizhou University, Taizhou 318000, China.}
	\thanks{Cong Bai is with the College of Computer Science and Technology, Zhejiang University of Technology, Hangzhou 310023, China.}
	\thanks{Corresponding authors: Zhen Zhang and Shiqing Zhang (e-mail: zhangzhen-911@tzc.edu.cn; zhangshiqing@tzc.edu.cn).}}

\maketitle
\begin{abstract}
Accurate short-term PM$_{2.5}$ forecasting is important for public health protection, air-quality early warning, and urban environmental management. However, PM$_{2.5}$ variation is driven by multiple coupled factors, including stable periodic changes induced by human activities and meteorological regularity, station-specific short-term concentration evolution, and meteorology-driven pollutant dispersion among monitoring stations. Existing spatio-temporal forecasting methods may capture station relationships to some extent, but distance-only, correlation-based, or purely adaptive graphs are often insufficient to comprehensively represent these heterogeneous factors, especially wind-direction-dependent pollutant transport. To address this problem, we propose a Multi-View Geo-Wind Guided KAN model for PM$_{2.5}$ forecasting, named \textbf{MVG-KAN}, which models station-level PM$_{2.5}$ evolution from three complementary views: local periodic regularity, station-wise residual temporal dynamics, and meteorological-environment-guided spatial dispersion. Specifically, the periodic-residual forecasting backbone first separates stable daily and weekly patterns from non-periodic residual variations. A Geo-Wind Graph is constructed by combining geographic distance decay with wind-direction- and wind-speed-aware transport, providing a lightweight physically motivated directed spatial prior for residual propagation among stations. In addition, a temporal Kolmogorov-Arnold network (TKAN) residual head is then introduced to learn station-wise nonlinear autoregressive correction from de-periodized PM$_{2.5}$ residuals and historical multi-pollutant sequences, thereby enhancing the modeling of local residual inertia and pollutant co-variation.  Experiments on the Beijing PM$_{2.5}$ dataset show that MVG-KAN achieves an MAE of 14.09 and an RMSE of 21.40. Ablation studies further validate the complementary effects of TKAN and Geo-Wind Graph, indicating that multi-view modeling is effective for station-level PM$_{2.5}$ forecasting.
\end{abstract}

\section{Introduction}

Fine particulate matter with an aerodynamic diameter below $2.5~\mu\mathrm{m}$, namely PM$_{2.5}$, is one of the most harmful urban air pollutants. High PM$_{2.5}$ concentration is closely associated with respiratory and cardiovascular diseases, reduced atmospheric visibility, and degraded urban environmental quality \cite{pope2006health,brook2010cardiovascular,hyslop2009visibility,lelieveld2015mortality}. Accurate PM$_{2.5}$ forecasting is therefore important for public health protection, air-quality early warning, exposure risk assessment, and urban environmental management \cite{zhang2012realtimeaqf,bai2018airpollutionforecasts}. Reliable forecasts over the next several hours or days can support public advisories, temporary emission-control planning, traffic and industrial regulation, and emergency response during heavy pollution episodes.

\begin{figure}[!t]
	\centering
	\includegraphics[width=0.48\textwidth]{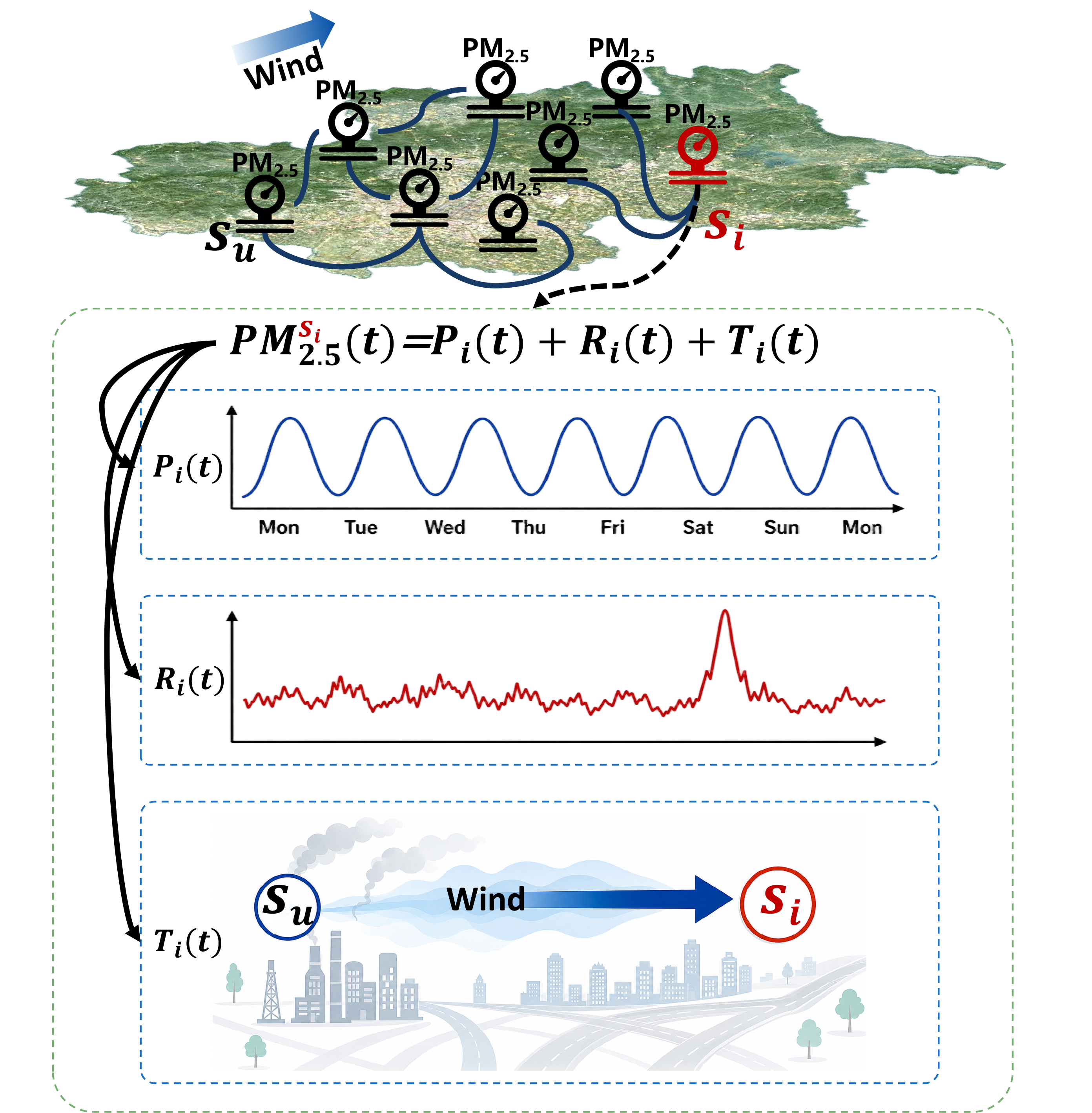}
	\caption{Motivation of MVG-KAN for PM$_{2.5}$ forecasting. PM$_{2.5}$ evolution at a target station $S_i$ is modeled as multi-view of local periodic pattern $P_i(t)$, local residual dynamics $R_i(t)$, and wind-driven spatial transport $T_i(t)$. The periodic component describes regular daily/weekly variations, the residual component represents local non-periodic fluctuations and pollution episodes, and the transport component captures directed pollutant propagation from upwind/source stations to $S_i$ under wind fields. This decomposition motivates the proposed framework to jointly model temporal regularity, station-wise residual correction, and physically informed geo-wind spatial transport, rather than relying only on temporal history or generic distance/correlation/adaptive graphs.}
	\label{fig:1}
\end{figure}

To achieve accurate PM$_{2.5}$ forecasting, a variety of approaches have been developed over the past decades. Early studies mainly relied on traditional statistical methods, such as autoregressive integrated moving average (ARIMA), regression models, and other time-series forecasting techniques \cite{brockwell2002introduction}. These methods are computationally efficient and provide relatively good interpretability, making them suitable for capturing simple temporal trends and periodic fluctuations. However, they usually rely on linear assumptions and handcrafted feature engineering, which limits their ability to model the highly nonlinear and complex interactions among pollutants, meteorological variables, and emission sources \cite{zheng2013uair,li2017lstm_air,qi2018deepairlearning}. As PM$_{2.5}$ formation and evolution involve intricate nonlinear processes, traditional statistical methods often struggle to maintain forecasting accuracy under rapidly changing environmental conditions \cite{bai2018airpollutionforecasts}.

To overcome these limitations, deep learning methods have been increasingly adopted for PM$_{2.5}$ forecasting. Models such as recurrent neural networks (RNNs) \cite{elman1990finding}, long short-term memory networks (LSTMs)\cite{hochreiter1997lstm}, gated recurrent units (GRUs) \cite{cho2014gru}, temporal convolutional networks (TCNs) \cite{bai2018tcn}, and Transformer-based architectures have demonstrated strong capabilities in learning nonlinear temporal dependencies directly from historical observations \cite{lim2021tft,zhou2021informer,liu2024itransformer,11345165}. By leveraging large-scale monitoring data, these methods can automatically extract complex temporal patterns without requiring explicit assumptions about the underlying data distribution. Nevertheless, most early deep learning approaches primarily focus on temporal modeling at individual stations and often overlook the spatial interactions among monitoring locations. Since air pollutants can be transported across regions through atmospheric processes, ignoring spatial dependencies may lead to incomplete representations of PM$_{2.5}$ dynamics.

To address this issue, spatio-temporal forecasting models have been proposed to jointly model temporal evolution and spatial interactions \cite{liang2018geoman,yi2018deepair,qi2019gclstm,deeppm25}. These methods integrate temporal neural networks with spatial modules to capture correlations among monitoring stations, thereby improving forecasting performance in regional air-quality prediction tasks. Compared with purely temporal models, spatio-temporal approaches can better characterize pollutant propagation and regional co-variation patterns. However, many existing spatio-temporal methods represent spatial relationships using predefined grids or generic spatial operators, which may not adequately describe the irregular distribution of monitoring stations and the heterogeneous interactions among them. Consequently, their ability to model realistic inter-station dependencies remains limited.

Graph-based methods provide a natural solution to this challenge by representing monitoring stations as graph nodes and station relationships as graph edges. Graph neural networks (GNNs) \cite{kipf2017gcn,velickovic2018gat} and spatio-temporal graph neural networks (STGNNs) \cite{qi2019gclstm} have achieved remarkable success in air-quality forecasting because they can flexibly model irregular spatial structures and learn complex dependencies among stations \cite{li2018dcrnn,yu2018stgcn,wang2020pm25gnn,wu2019graphwavenet}. Existing graph construction strategies generally rely on geographic distance, statistical correlation, adaptive graph learning, or combinations thereof. These approaches effectively capture certain aspects of spatial relationships and have significantly advanced PM$_{2.5}$ forecasting performance.

Despite these advances, existing graph-based PM$_{2.5}$ forecasting methods still exhibit several limitations. First, many methods focus primarily on learning spatial dependencies while paying insufficient attention to the intrinsic periodic regularity of PM$_{2.5}$ evolution \cite{wu2021autoformer,zeng2023dlinear,shao2026hyperd}. In practice, human activities, traffic intensity, industrial production, and meteorological conditions often follow daily and weekly cycles, resulting in stable periodic patterns in PM$_{2.5}$ concentrations. Explicitly modeling such periodic regularity can provide strong predictive signals and improve forecasting robustness.

Second, after periodic patterns are removed, the remaining PM$_{2.5}$ residuals should not be regarded as random patterns. These residuals often contain station-specific temporal dynamics arising from pollutant accumulation, slow dissipation, local emissions, abrupt environmental changes, and interactions among multiple pollutants. Existing graph-based methods typically emphasize global spatio-temporal representations but may underexploit the nonlinear autoregressive characteristics embedded in station-wise residual sequences.

Third, pollutant transport is fundamentally influenced by meteorological conditions, especially wind fields. Pollutants emitted at an upwind station may significantly affect downwind locations, whereas geographically nearby stations may exert limited influence if they are not aligned with the prevailing wind direction. However, many existing graph construction strategies rely solely on geographic distance, historical correlation, or adaptive learning mechanisms. Such graphs may capture statistical dependencies but often lack explicit physical guidance regarding meteorology-driven pollutant dispersion \cite{yang2019dwfd,zhou2021ddstgcn,xiao2022dpddgcn}.

As illustrated in Fig.~\ref{fig:1}, these observations suggest that accurate PM$_{2.5}$ forecasting should simultaneously consider three complementary views: periodic regularity, station-specific residual dynamics, and meteorology-driven spatial transport. From this perspective, the evolution of PM$_{2.5}$ concentration at station $i$ can be conceptually expressed as

\begin{equation}
	PM_{2.5}^{i}(t)=P_i(t)+R_i(t)+T_i(t),
\end{equation}

where $P_i(t)$ denotes the local periodic component, $R_i(t)$ denotes the station-wise residual temporal component, and $T_i(t)$ denotes the meteorology-driven transport component. This formulation is not intended as a strict physical decomposition of atmospheric processes. Instead, it provides a forecasting-oriented multi-view perspective that highlights three major sources of predictability in PM$_{2.5}$ evolution.

The periodic component captures stable temporal regularity induced by recurring human activities and environmental cycles. The residual component captures local nonlinear dynamics that remain predictable after periodicity is removed. The transport component captures inter-station influence caused by pollutant dispersion under changing meteorological conditions. These three views are complementary rather than redundant. Ignoring any one of them may result in an incomplete understanding of PM$_{2.5}$ dynamics and consequently limit forecasting performance.

Motivated by this observation, we propose \textbf{MVG-KAN}, a Multi-View Geo-Wind Guided KAN framework for PM$_{2.5}$ forecasting. The proposed framework explicitly models PM$_{2.5}$ evolution from the three perspectives described above. Specifically, we adopt a periodic-residual forecasting paradigm \cite{shao2026hyperd} as the backbone to separate stable periodic patterns from non-periodic variations. Building upon this foundation, we first develop a Geo-Wind Graph to model meteorology-driven pollutant dispersion by integrating geographic distance decay, wind direction, and wind speed into a unified directed graph structure. Unlike conventional distance-based or correlation-based graphs, the proposed Geo-Wind Graph explicitly incorporates physically meaningful transport information, allowing the model to distinguish between upwind and downwind relationships and better characterize pollutant propagation among monitoring stations.

Furthermore, we introduce a Temporal Kolmogorov--Arnold Network (TKAN) residual head to model station-wise nonlinear residual dynamics. Instead of directly learning from raw PM$_{2.5}$ sequences, TKAN operates on de-periodized residuals together with historical multi-pollutant observations, enabling it to focus on local residual inertia, abrupt changes, and pollutant co-variation patterns.

The periodic-residual backbone, Geo-Wind Graph spatial modeling, and TKAN residual temporal modeling jointly operationalize the proposed multi-view forecasting perspective. The periodic branch captures stable regularity, the Geo-Wind Graph captures meteorology-guided spatial transport, and TKAN captures station-specific residual evolution. Together, they provide a more comprehensive representation of PM$_{2.5}$ dynamics than existing approaches that focus on only one or two aspects of the forecasting problem.

The main contributions of this paper are summarized as follows:

\begin{enumerate}
	\item To address the insufficient decomposition of PM$_{2.5}$ evolution mechanisms in existing methods, we propose a Multi-View Geo-Wind Guided KAN framework (\textbf{MVG-KAN}), which jointly models local periodic regularity, station-wise residual temporal dynamics, and meteorology-driven spatial dispersion.
	
	\item To overcome the limitation of distance-only, correlation-based, or purely adaptive spatial graphs, we develop a Geo-Wind Graph that integrates geographic distance decay, wind direction, and wind speed for directed pollutant dispersion modeling.
	
	\item To better capture non-periodic local fluctuations after removing stable periodic patterns, we introduce a TKAN residual head that learns nonlinear autoregressive correction from de-periodized PM$_{2.5}$ residuals and historical multi-pollutant sequences.
	
	\item Extensive experiments on the Beijing PM$_{2.5}$ dataset demonstrate the effectiveness of MVG-KAN. The proposed model achieves an MAE of $14.09$ and an RMSE of $21.40$, yielding the best average MAE and RMSE among the compared deep learning methods. Further ablation studies verify the complementary roles of TKAN and the Geo-Wind Graph, where TKAN improves local residual temporal correction and the Geo-Wind Graph provides physically meaningful spatial enhancement.
\end{enumerate}

\section{Related Work}

\subsection{PM$_{2.5}$ and Air Quality Forecasting}

PM$_{2.5}$ forecasting has been studied through statistical time-series modeling, atmospheric transport modeling, remote-sensing estimation, classical machine learning, and deep learning \cite{zhang2012realtimeaqf,bai2018airpollutionforecasts}. Statistical and classical models provide interpretable temporal baselines, but their linear assumptions and handcrafted features are often inadequate for nonlinear emission effects, meteorological dispersion, and multi-station dependencies \cite{zheng2013uair,li2017lstm_air}. Chemical transport and atmospheric dispersion systems, such as HYSPLIT, explicitly represent physical processes and trajectory information \cite{stein2015hysplit}. Remote-sensing-based PM$_{2.5}$ estimation studies further combine aerosol observations, chemical transport modeling, meteorological information, and ground monitoring to characterize spatial pollution fields \cite{geng2015rsepm25,wei2019strfpm25,chen2023tgrspm25}. These studies provide strong environmental motivation, but they are often designed for physical simulation or spatial estimation rather than lightweight station-level multi-step forecasting.

Data-driven air-quality models reduce the dependence on explicit physical simulation by learning from observed pollutant, meteorological, and station-context variables. Hybrid trajectory-wavelet models, urban air-quality inference, DeepAir, distributed fusion networks, GeoMAN, and recent DeepPM$_{2.5}$ forecasting systems all demonstrate the value of heterogeneous predictors for fine-grained PM$_{2.5}$ prediction \cite{feng2015trajectorywavelet,zheng2013uair,qi2018deepairlearning,yi2018deepair,liang2018geoman,deeppm25}. However, richer inputs alone do not guarantee that a model distinguishes directional pollutant transport from local temporal evolution. This motivates a station-level forecasting framework that explicitly separates local periodic regularity, local residual dynamics, and wind-guided spatial transport.

\subsection{Spatio-Temporal Graph Learning for Environmental Prediction}

Multi-station air-quality forecasting is naturally related to spatio-temporal graph learning, where monitoring stations are graph nodes and edges encode spatial dependence, transport pathways, or statistical similarity. Graph convolution, graph attention, diffusion-style propagation, and spatio-temporal graph convolution provide common tools for irregular sensor networks \cite{kipf2017gcn,velickovic2018gat,li2018dcrnn,yu2018stgcn}. In PM$_{2.5}$ forecasting, graph convolution with recurrent temporal modeling and domain-knowledge-enhanced graph construction have been used to capture station relationships and environmental priors \cite{qi2019gclstm,wang2020pm25gnn}. Adaptive graph learning, represented by Graph WaveNet, further allows latent dependencies to be inferred when fixed adjacency information is incomplete \cite{wu2019graphwavenet}.

Despite these advances, common distance-based, correlation-based, and adaptive graphs may not sufficiently reflect pollutant transport. PM$_{2.5}$ movement depends not only on proximity or historical similarity, but also on wind direction, wind speed, and upwind/downwind geometry. A nearby station may be weakly related if it is not aligned with transport direction, whereas an upwind station can be influential even when it is not the nearest neighbor. Wind-aware forecasting studies have considered dynamic wind-field distance, wind-conditioned graph convolution, and directed dynamic graph structures \cite{yang2019dwfd,zhou2021ddstgcn,xiao2022dpddgcn}. These works indicate that wind variables should shape spatial propagation rather than serve only as auxiliary node features.

The present work follows this transport-aware perspective. Instead of relying on a generic station graph, it introduces a Geo-Wind Graph to provide a lightweight directed spatial prior based on geographic proximity and wind-driven transport tendency. The graph is intended to guide residual spatial propagation in a physically motivated manner, not to replace chemical transport or atmospheric dispersion models.

\subsection{Decomposition and Residual Modeling for Time Series Forecasting}

Time-series decomposition is widely used when observations contain both regular temporal structure and irregular fluctuations. Environmental sequences often exhibit daily or weekly cycles caused by human activity, meteorological variation, and regional background conditions. In PM$_{2.5}$ forecasting, trajectory-based geographic features and wavelet transformation show that separating temporal scales can help model pollution episodes \cite{feng2015trajectorywavelet}. In broader time-series forecasting, decomposition-based Transformer models and simple linear temporal mappings also demonstrate that explicit seasonal or residual representations can be competitive \cite{wu2021autoformer,zeng2023dlinear}. Recent periodic-residual frameworks, such as HyperD, further decouple stable periodic patterns from non-periodic residual components \cite{shao2026hyperd}.

For station-level PM$_{2.5}$ prediction, however, the residual after removing periodicity should not be treated as unstructured noise. It may retain pollutant persistence, accumulation, slow dissipation, local emission changes, and co-variation among PM$_{2.5}$, PM$_{10}$, NO$_2$, CO, SO$_2$, O$_3$, and AQI. Multi-source air-quality models have shown that pollutant and meteorological variables provide complementary evidence for future concentration changes \cite{qi2018deepairlearning,yi2018deepair}. Nevertheless, existing decomposition and residual branches often emphasize global temporal representations or frequency-domain alignment, while station-wise nonlinear residual inertia and multi-pollutant co-variation remain under-modeled.

The TKAN residual temporal head is introduced to address this local residual modeling gap. Motivated by Kolmogorov--Arnold networks \cite{liu2024kan}, it focuses on station-wise residual correction from de-periodized PM$_{2.5}$ and recent multi-pollutant histories. It is complementary to the Geo-Wind Graph: TKAN targets local temporal residual dynamics, whereas the graph module targets wind-guided inter-station transport.
\begin{figure*}[!t]
	\centering
	\includegraphics[width=0.98\textwidth]{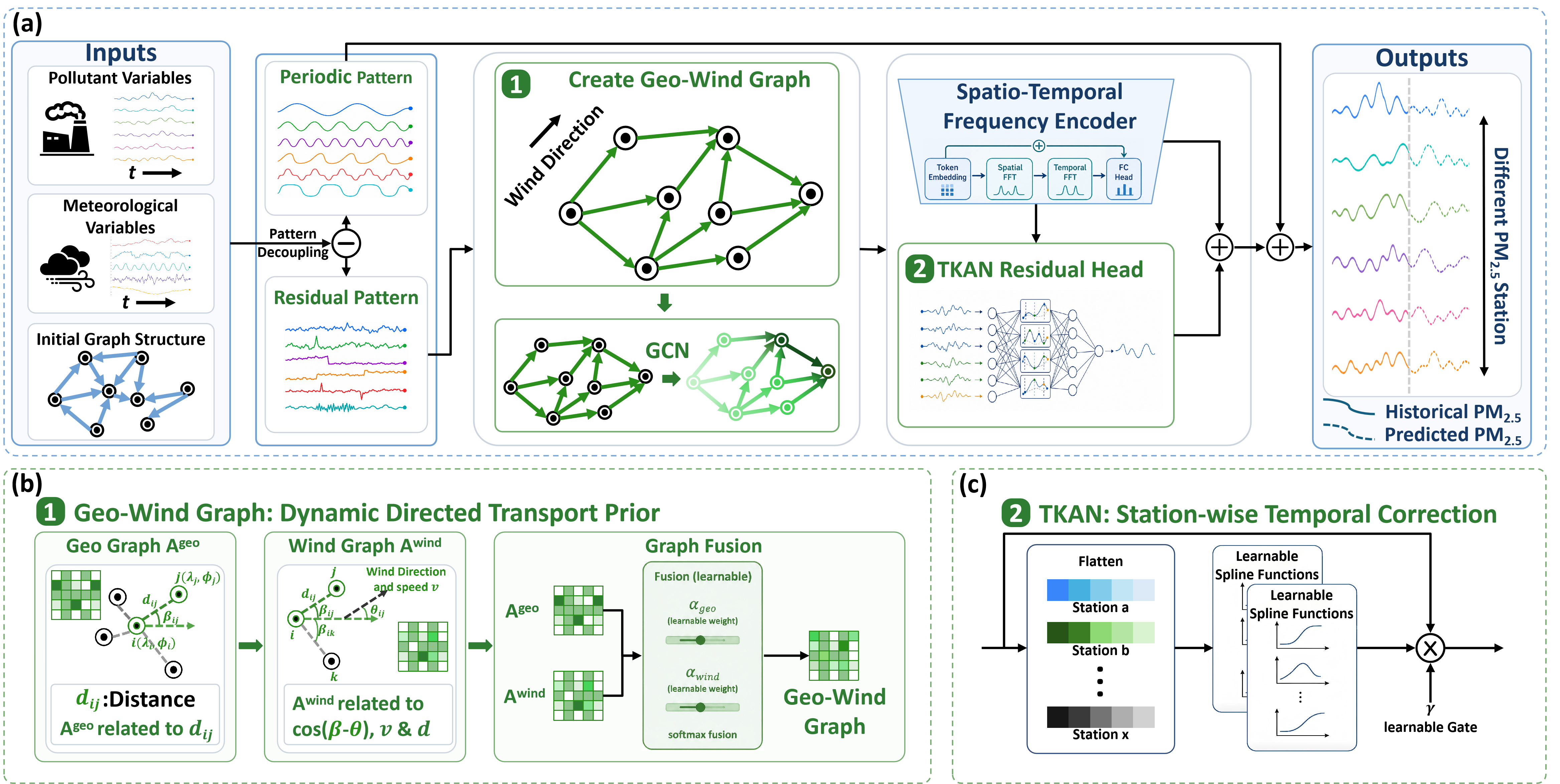}
	\caption{Overall framework of the geo-wind guided KAN residual temporal learning model. (a) Overall framework. The input pollutant and meteorological sequences are decomposed into periodic and residual patterns. The periodic branch models regular temporal variations, while the residual branch is refined by the proposed Geo-Wind Graph and TKAN residual head to represent wind-driven spatial transport $T_i(t)$ and local residual dynamics $R_i(t)$, respectively. The final PM$_{2.5}$ forecast is reconstructed by combining the periodic component with the refined residual component.
		(b) Geo-Wind Graph module. A dynamic directed transport graph is constructed by combining geographic proximity with wind-direction- and wind-speed-aware modulation, providing a physically motivated spatial prior for pollutant propagation.
		(c) TKAN residual head. A station-wise temporal KAN maps de-periodized residual and pollutant histories to nonlinear residual corrections, strengthening local temporal residual modeling after periodic effects are removed.}
	\label{fig:overall_framework}
\end{figure*}
\section{Method}

\subsection{Problem Formulation}

Let $\mathbf{X}\in\mathbb{R}^{T\times N\times C}$ denote a multi-station and multi-variate historical observation window, where $T$ is the input sequence length, $N$ is the number of PM$_{2.5}$ monitoring stations, and $C$ is the number of value features. In the current implementation, $T=48$, $N=39$, and $C=12$ for the final full-feature configuration. The value features are AQI, CO, NO$_2$, O$_3$, PM$_{10}$, SO$_2$, dew-point temperature, ground pressure, temperature, wind direction, wind speed, and PM$_{2.5}$. In addition to these value features, the model receives hour-of-day and day-of-week indicators for periodic indexing. The target is to predict the future PM$_{2.5}$ sequence
\begin{equation}
\mathbf{Y}\in\mathbb{R}^{H\times N},
\end{equation}
where $H$ is the prediction horizon. The implemented setting uses $H=48$. For a mini-batch, we write the historical value tensor as $\mathbf{X}\in\mathbb{R}^{B\times T\times N\times C}$ and the PM$_{2.5}$ target as $\mathbf{Y}\in\mathbb{R}^{B\times H\times N}$. The forecasting objective is short-term station-level PM$_{2.5}$ prediction; the output channel is PM$_{2.5}$ only.

\subsection{Overall Framework}

The proposed framework is built on a periodic-residual forecasting backbone~\cite{shao2026hyperd}. The backbone contains a periodic representation branch and a residual representation branch. The periodic branch uses temporal indices to obtain an input-period periodic pattern $\mathbf{S}^{in}\in\mathbb{R}^{B\times T\times N}$ and a future periodic pattern $\mathbf{S}^{out}\in\mathbb{R}^{B\times H\times N}$. Let $\mathbf{X}^{p}\in\mathbb{R}^{B\times T\times N}$ denote the PM$_{2.5}$ channel selected from $\mathbf{X}$. The input residual is
\begin{equation}
\mathbf{R}^{in}=\mathbf{X}^{p}-\mathbf{S}^{in}.
\label{eq:residual_in}
\end{equation}
The residual branch models the non-periodic part of the sequence. In our framework, two PM$_{2.5}$-oriented modules are added to this branch, as shown in Fig.~\ref{fig:overall_framework}.(a). First, a Geo-Wind Graph constructs a dynamic directed adjacency tensor $\mathbf{A}_{GW}\in\mathbb{R}^{B\times T\times N\times N}$ from station coordinates and wind variables. This module models wind-driven spatial transport of non-periodic PM$_{2.5}$ residual disturbances across stations. Second, a TKAN residual head is applied after the residual output to provide station-wise residual correction from de-periodized PM$_{2.5}$ and historical pollutant variables.

Let $\widehat{\mathbf{R}}\in\mathbb{R}^{B\times H\times N}$ denote the final residual prediction after Geo-Wind Graph enhancement and TKAN residual correction. Specifically, the Geo-Wind Graph provides a dynamic spatial correction for the transport-related residual component $\mathbf{T}_i(t)$, while the TKAN residual head captures station-wise nonlinear temporal residual dynamics $\mathbf{R}_i(t)$. The final PM$_{2.5}$ prediction $\widehat{\mathbf{Y}}$ is obtained by adding the future periodic component $\mathbf{S}^{out}$ and the final residual prediction:
\begin{equation}
	\widehat{\mathbf{Y}}=\mathbf{S}^{out}+\widehat{\mathbf{R}}.
	\label{eq:final_prediction}
\end{equation}
Here, $\mathbf{S}^{out}\in\mathbb{R}^{B\times H\times N}$ represents the predicted daily-weekly periodic component, and $\widehat{\mathbf{R}}$ represents the non-periodic residual component refined by spatial transport modeling and local temporal correction. Therefore, the final prediction decomposes PM$_{2.5}$ into an explicit periodic part and a learned residual part.

\subsection{Geo-Wind Guided Graph Construction}

\begin{figure}[!t]
\centering
\includegraphics[width=\columnwidth]{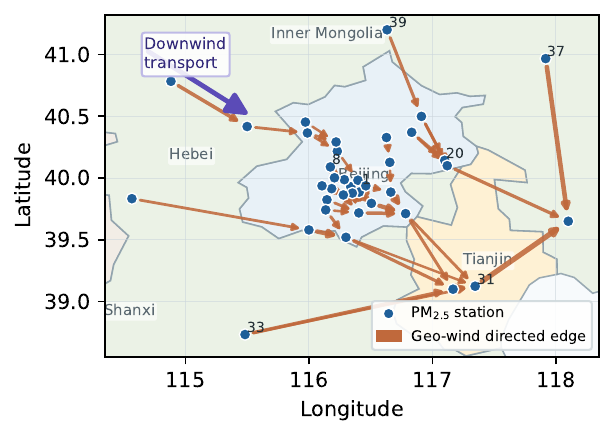}
\caption{Geo-Wind Graph illustration. Blue nodes denote the 39 PM$_{2.5}$ monitoring stations, and orange arrows show representative directed transport edges constructed from geographic proximity and downwind alignment. The upper-left arrow indicates the downwind transport direction used for visualization.}
\label{fig:geo_wind_graph}
\end{figure}

PM$_{2.5}$ dependence among monitoring stations is affected by both geographic proximity and wind-driven transport. As visualized in Fig.~\ref{fig:geo_wind_graph} and Fig.~\ref{fig:overall_framework}.(b), the Beijing monitoring stations form an irregular spatial network in which plausible transport edges should depend on station locations and downwind alignment rather than distance alone. The graph of PM$_{2.5}$ should therefore not be treated as a simple static distance graph or a purely data-driven correlation graph, but as a physically informed transport graph that explicitly models wind-driven pollutant propagation between monitoring stations. The implemented Geo-Wind Graph constructs a directed adjacency tensor in which $\mathbf{A}_{GW}[b,t,i,j]$ represents the strength of pollutant propagation from source station $i$ to target station $j$ at time $t$ in batch sample $b$.

\subsubsection{Geographic Proximity}

For each station $i$, let $(\lambda_i,\phi_i)$ denote longitude and latitude in radians. We precompute the haversine distance between stations $i$ and $j$ as
\begin{equation}
a_{ij}=\sin^2\left(\frac{\phi_j-\phi_i}{2}\right)
 +\cos\phi_i\cos\phi_j\sin^2\left(\frac{\lambda_j-\lambda_i}{2}\right),
\end{equation}
\begin{equation}
d_{ij}=2R_e\operatorname{atan2}\left(\sqrt{a_{ij}},\sqrt{1-a_{ij}}\right),
\label{eq:haversine}
\end{equation}
where $R_e=6371$ km. It also precomputes the bearing $\beta_{ij}$ from station $i$ to station $j$:
\begin{equation}
\begin{aligned}
\beta_{ij}=\operatorname{atan2}\big(&
\sin(\lambda_j-\lambda_i)\cos\phi_j,\\
&\cos\phi_i\sin\phi_j
-\sin\phi_i\cos\phi_j\cos(\lambda_j-\lambda_i)
\big),
\end{aligned}
\label{eq:bearing}
\end{equation}
wrapped to $[0,2\pi)$. The geographic graph is initialized by an exponential distance kernel,
\begin{equation}
\widetilde{A}^{geo}_{ij}=
\begin{cases}
\exp(-d_{ij}/\tau_d), & i\ne j,\\
0, & i=j,
\end{cases}
\label{eq:geo_kernel}
\end{equation}
where the final configuration uses $\tau_d=50.0$. A top-$k$ operation keeps the largest $k$ outgoing neighbors for each source station, followed by row normalization:
\begin{equation}
A^{geo}_{ij}=
\frac{M_{ij}\widetilde{A}^{geo}_{ij}}
{\sum_{\ell=1}^{N}M_{i\ell}\widetilde{A}^{geo}_{i\ell}+\epsilon},
\label{eq:topk_norm}
\end{equation}
where $M_{ij}\in\{0,1\}$ is the top-$k$ mask, $k=8$ in the final configuration, and $\epsilon$ is a small constant for numerical stability.

\subsubsection{Wind Graph}

The wind graph uses wind direction and wind speed from each source station, as shown in Fig. \ref{fig:overall_framework}.(b). For degree-valued input, the downwind transport direction $\theta^{down}_{bti}$ is
\begin{equation}
\theta^{down}_{bti}=\operatorname{deg2rad}\left((\theta^{wind}_{bti}+180)\bmod 360\right),
\label{eq:downwind}
\end{equation}
where $\theta^{wind}_{bti}$ is the recorded wind direction.

The directional alignment between the downwind direction at source station $i$ and the bearing from $i$ to $j$ is
\begin{equation}
D_{btij}=\left[\operatorname{ReLU}\left(\cos(\beta_{ij}-\theta^{down}_{bti})\right)\right]^{p_d},
\label{eq:direction_alignment}
\end{equation}
where $D_{btij}$ measures the likelihood that pollutants emitted or accumulated at station $i$ can be transported toward station $j$ under the current wind conditions. It acts as a directional transport gate that favors downwind connections and suppresses upwind or crosswind interactions, $p_d=2.0$ in the final configuration. A wind-speed gate is computed as
\begin{equation}
G_{bti}=\sigma\left(\frac{v_{bti}-v_0}{s_v+\epsilon}\right),
\label{eq:speed_gate}
\end{equation}
where $v_{bti}$ is the wind speed, $v_0=1.0$, and $s_v=0.5$ in the final configuration. The $\sigma()$ gate converts wind speed into a smooth transport-strength coefficient in the range $(0,1)$. When the wind speed is lower than $v_0$, the gate value is suppressed, indicating limited pollutant transport. As wind speed increases beyond $v_0$, the gate gradually approaches 1, allowing stronger wind-driven spatial interactions. For travel-time decay, the code first converts the wind speed into an effective transport velocity and enforces a lower bound to avoid unrealistically large travel times under near-calm conditions:
\begin{equation}
	\widetilde{v}_{bti}=\max(v_{bti},v_{min})\times 3.6,
\end{equation}
where $v_{min}=0.2$. The factor $3.6$ converts wind speed from m/s to km/h, making it consistent with the distance unit used in the graph construction.

The estimated travel time required for pollutants to move from station $i$ to station $j$ is then approximated as
\begin{equation}
	T^{travel}_{btij}=\frac{d_{ij}}{\widetilde{v}_{bti}+\epsilon},
\end{equation}
where $d_{ij}$ denotes the geographic distance between the two stations. Larger distances or weaker winds result in longer travel times.

To model the gradual attenuation of transport influence over time, an exponential decay function is applied:
\begin{equation}
	Q_{btij}=\exp\left(-\frac{T^{travel}_{btij}}{\tau_t+\epsilon}\right),
	\label{eq:travel_decay}
\end{equation}
where $\tau_t=6.0$ controls the characteristic transport timescale. The decay factor $Q_{btij}\in(0,1]$ assigns larger weights to station pairs that can be connected more rapidly by the current wind field, while suppressing interactions requiring excessively long transport times. Consequently, nearby stations under strong winds retain strong connections, whereas distant stations or stations connected through weak winds receive smaller weights.

The wind-directed graph is then constructed by combining geographic proximity, directional consistency, wind-speed strength, and travel-time feasibility:
\begin{equation}
	\widetilde{A}^{wind}_{btij}
	=A^{geo}_{ij}\,D_{btij}\,G_{bti}\,Q_{btij},
	\label{eq:wind_graph_raw}
\end{equation}
where $A^{geo}_{ij}$ provides the distance-based spatial prior, $D_{btij}$ favors station pairs aligned with the downwind direction, $G_{bti}$ controls whether the wind at the source station is strong enough to support transport, and $Q_{btij}$ suppresses connections whose estimated travel time is too long.

The raw wind-directed graph is further sparsified and normalized using the same top-$k$ row normalization as in \eqref{eq:topk_norm}. This ensures that each station only aggregates information from the most physically plausible upstream or downstream neighbors.

Since both the downwind direction $\theta^{down}_{bti}$ and wind speed $v_{bti}$ depend on the source station $i$ and time step $t$, the resulting wind graph is dynamic and directed. In general, the transport influence from station $i$ to station $j$ differs from that from station $j$ to station $i$, i.e.,
\[
A^{wind}_{btij} \neq A^{wind}_{btji}.
\]
This asymmetry is consistent with wind-driven pollutant transport, because pollutants are more likely to propagate along the downwind direction rather than equally in both directions.

\subsubsection{Graph Fusion}

The final confirmed Geo-Wind configuration uses two graphs, $\mathbf{A}^{geo}$ and $\mathbf{A}^{wind}$. They are fused by learnable softmax weights:
\begin{equation}
\alpha_m=\frac{\exp(l_m)}{\sum_{q}\exp(l_q)},\qquad
\widetilde{\mathbf{A}}_{GW}=\sum_{m\in\{geo,wind\}}\alpha_m\mathbf{A}^{m},
\label{eq:graph_fusion}
\end{equation}
where $l_m$ are learnable graph logits. The fused graph is again sparsified by top-$k$ row normalization, passed through dropout, and normalized once more. The final graph is $\mathbf{A}_{GW}\in\mathbb{R}^{B\times T\times N\times N}$.

\subsubsection{Graph Convolution Based on Geo-Wind Graph}

The Geo-Wind graph is inserted into the residual branch to provide wind-aware spatial correction before temporal residual forecasting. Given the historical PM$_{2.5}$ residual $\mathbf{R}^{in}\in\mathbb{R}^{B\times T\times N}$, which is calculated by equation.\ref{eq:residual_in}, we first normalize and project it into a hidden feature space:
\begin{equation}
	\mathbf{H}^{(0)}
	=
	f_{\mathrm{enc}}
	\left(
	\operatorname{Norm}
	\left(
	\mathbf{R}^{in}
	\right)
	\right),
\end{equation}
where $\mathbf{H}^{(0)}\in\mathbb{R}^{B\times T\times N\times C}$, $C$ denotes the hidden dimension, and $f_{\mathrm{enc}}$ dotes an MLP encoder. In the final configuration, only the PM$_{2.5}$ residual is used as the node feature for graph convolution, while meteorological variables are used to construct the dynamic Geo-Wind adjacency $\mathbf{A}_{GW}$.

For the directed adjacency $\mathbf{A}_{GW}$, the entry $A_{GW,btij}$ denotes the transport weight from source station $i$ to target station $j$ at batch $b$ and time step $t$. A larger $A_{GW,btij}$ means that the residual information at station $i$ is more likely to influence station $j$ under the current wind field. The $k$-th order forward diffusion is defined as
\begin{equation}
	\mathbf{H}^{(k)}_{f,btj}
	=
	\sum_{i=1}^{N}
	A_{GW,btij}
	\mathbf{H}^{(k-1)}_{f,bti},
	\label{eq:forward_diffusion}
\end{equation}
where $\mathbf{H}^{(k)}_{f,btj}$ is the forward aggregated representation of target station $j$. This operation updates station $j$ by collecting residual features from source stations according to their wind-directed transport weights. Therefore, stations that are geographically close, aligned with the downwind direction, and reachable within a reasonable travel time contribute more strongly to the representation of station $j$.

Since the Geo-Wind graph is directed, the transport influence from $i$ to $j$ is generally different from that from $j$ to $i$. To model reverse directional dependencies, backward diffusion is computed using the transposed adjacency $\mathbf{A}_{GW}^{\top}$:
\begin{equation}
	\mathbf{H}^{(k)}_{b,bti}
	=
	\sum_{j=1}^{N}
	A_{GW,btij}
	\mathbf{H}^{(k-1)}_{b,btj}.
	\label{eq:backward_diffusion}
\end{equation}
Thus, forward diffusion captures information propagated along the wind-directed graph, while backward diffusion $\mathbf{H}^{(k)}_{b,bti}$ captures information propagated along the reverse direction.

With graph order $K=2$, the graph convolution considers both one-hop and two-hop spatial diffusion. The original residual feature, two forward diffusion supports, and two backward diffusion supports are concatenated as
\begin{equation}
	\mathbf{Z}_{bt}
	=
	\operatorname{Concat}
	\left(
	\mathbf{H}^{(0)}_{bt},
	\mathbf{H}^{(1)}_{f,bt},
	\mathbf{H}^{(2)}_{f,bt},
	\mathbf{H}^{(1)}_{b,bt},
	\mathbf{H}^{(2)}_{b,bt}
	\right).
\end{equation}
The first-order supports capture direct wind-driven interactions between stations, while the second-order supports capture indirect interactions through intermediate stations. The concatenated representation is then passed through a linear projection, GELU activation, dropout, and a residual connection to obtain the graph-enhanced hidden $\mathbf{H}^{graph}$ representation:
\begin{equation}
	\mathbf{H}^{graph}
	=
	\mathbf{H}^{(0)}
	+
	\operatorname{Dropout}
	\left(
	\operatorname{GELU}
	\left(
	\operatorname{Linear}
	\left(
	\mathbf{Z}
	\right)
	\right)
	\right).
\end{equation}

Finally, the graph-enhanced hidden representation is decoded back to the residual space:
\begin{equation}
	\Delta\mathbf{R}^{graph}
	=
	f_{\mathrm{dec}}
	\left(
	\mathbf{H}^{graph}
	\right),
\end{equation}
where $\Delta\mathbf{R}^{graph}\in\mathbb{R}^{B\times T\times N}$ and $f_{\mathrm{dec}}$ dotes an MLP decoder. This decoded graph output is used as a correction to the historical residual input:
\begin{equation}
	\widetilde{\mathbf{R}}^{in}
	=
	\mathbf{R}^{in}
	+
	\gamma_g
	\Delta\mathbf{R}^{graph},
	\label{eq:graph_residual_input}
\end{equation}
where $\gamma_g$ is a learnable gate initialized to $0.0$ in the final configuration. This initialization makes the model start from the original residual branch and gradually learn whether and how strongly the wind-aware graph correction should be used.

The graph-enhanced residual sequence $\widetilde{\mathbf{R}}^{in}$ is then fed into the residual temporal forecasting module, named Spatio-Temporal Frequency Encoder, inherited from the baseline branch. This module models temporal dependencies in the enhanced historical residual sequence and produces the preliminary future residual prediction $\mathbf{R}^{out}\in\mathbb{R}^{B\times H\times N}$. The output $\mathbf{R}^{out}$ is further combined with the station-wise TKAN residual correction introduced in the next subsection to obtain the final residual prediction $\widehat{\mathbf{R}}$. Therefore, the proposed graph convolution does not replace the temporal residual predictor; instead, it provides a dynamic wind-aware spatial correction before future residual forecasting, while TKAN further refines the predicted residuals through local nonlinear temporal correction.

\subsection{TKAN Residual Temporal Module}

The Geo-Wind Graph models wind-aware spatial propagation among monitoring stations, but it does not explicitly capture the nonlinear temporal evolution of the residual sequence at each individual station. After removing the daily and weekly periodic components, the remaining PM$_{2.5}$ residual may still contain short-term persistence, pollutant accumulation, slow dissipation, abrupt fluctuations, and co-variation with other pollutants. Therefore, we introduce a TKAN residual temporal module as a station-wise nonlinear correction head for future residual prediction, as shown in Fig. \ref{fig:overall_framework}.(c).

Let $\mathbf{X}^{tkan}\in\mathbb{R}^{B\times T\times N\times C_p}$ denote the input tensor of the TKAN module, where $C_p$ is the number of selected pollutant channels. In the final configuration, we use seven pollutant-related variables, including PM$_{2.5}$, AQI, PM$_{10}$, NO$_2$, CO, SO$_2$, and O$_3$, so $C_p=7$. To avoid duplicating the periodic modeling branch, the PM$_{2.5}$ channel is replaced by its de-periodized residual:
\begin{equation}
	r^{p}_{bti}=x^{p}_{bti}-S^{in}_{bti},
	\label{eq:tkan_residual_replacement}
\end{equation}
where $x^{p}_{bti}$ denotes the raw PM$_{2.5}$ observation at station $i$ and time step $t$, and $S^{in}_{bti}$ is the learned periodic component. This replacement forces TKAN to focus on residual temporal dynamics rather than relearning daily or weekly periodic patterns.

For each station $i$, the local historical sequence is processed independently. Specifically, the temporal window and pollutant channels are flattened into a station-wise vector:
\begin{equation}
	\mathbf{z}_{bi}
	=
	\operatorname{Flatten}
	\left(
	\mathbf{X}^{tkan}_{b,:,i,:}
	\right)
	\in\mathbb{R}^{T C_p}.
	\label{eq:tkan_flatten}
\end{equation}
In implementation, all station-wise vectors are reshaped from $[B,N,TC_p]$ to $[BN,TC_p]$ so that the same TKAN head is shared across stations. This design enables parameter sharing while preserving station-wise independent temporal modeling.

Following the KAN formulation~\cite{liu2024kan}, the two-layer TKAN head is formulated as
\begin{equation}
	\mathbf{h}_{bi}
	=
	\operatorname{KAN}_{1}
	\left(
	\mathbf{z}_{bi}; g,o
	\right),
	\label{eq:tkan_hidden}
\end{equation}
\begin{equation}
	\Delta\widehat{\mathbf{Y}}_{bi}
	=
	\operatorname{KAN}_{2}
	\left(
	\operatorname{Dropout}
	\left(
	\mathbf{h}_{bi}
	\right); g,o
	\right),
	\label{eq:tkan_output}
\end{equation}
where $g$ and $o$ denote the spline grid size and spline order, respectively. In the final configuration, $g=3$, $o=3$, and the hidden size is 32. The output $\Delta\widehat{\mathbf{Y}}_{bi}\in\mathbb{R}^{H}$ represents the TKAN-predicted residual correction for station $i$ over the future horizon. After reshaping, we obtain $\Delta\widehat{\mathbf{Y}}\in\mathbb{R}^{B\times H\times N}$.

The TKAN output is not treated as an independent forecast. Instead, it is used as a gated correction to the residual prediction produced by the main residual forecasting branch:
\begin{equation}
	\widehat{\mathbf{R}}
	=
	\mathbf{R}^{out}
	+
	\gamma_t
	\Delta\widehat{\mathbf{Y}},
	\label{eq:tkan_gate}
\end{equation}
where $\mathbf{R}^{out}\in\mathbb{R}^{B\times H\times N}$ denotes the residual output after Geo-Wind graph module and before TKAN correction, and $\gamma_t$ is a learnable gate initialized to $0.05$. This small initialization allows the model to start from the main residual predictor and gradually learn the contribution of TKAN during training. The corrected residual prediction $\widehat{\mathbf{R}}$ is then added to the future periodic component $\mathbf{S}^{out}$ to obtain the final PM$_{2.5}$ prediction, as defined in Equation.~\eqref{eq:final_prediction}.

Overall, TKAN serves as a station-wise nonlinear autoregressive residual correction module. It does not use graph information and does not perform spatial propagation. Therefore, it is complementary to the Geo-Wind Graph: the Geo-Wind Graph captures dynamic spatial transport under meteorological conditions, while TKAN captures local nonlinear temporal residual dynamics.

\subsection{Training Objective}

The model is trained using a horizon-aware masked mean absolute error (MAE) loss. Let $\mathcal{H}=\{1,3,6,12,24,48\}$ denote the set of evaluation horizons, and let $Y_{bhi}$ and $\widehat{Y}_{bhi}$ represent the ground-truth and predicted PM$_{2.5}$ values for sample $b$, horizon $h$, and station $i$, respectively. Since missing values are encoded as $0.0$ in the dataset, a binary mask $M_{bhi}$ is defined as
\begin{equation}
	M_{bhi}=
	\begin{cases}
		1, & Y_{bhi}\neq 0,\\
		0, & Y_{bhi}=0.
	\end{cases}
\end{equation}
The prediction loss is computed only on valid observations and selected evaluation horizons:
\begin{equation}
	\mathcal{L}_{pred}
	=
	\frac{\sum_{b,h\in\mathcal{H},i}M_{bhi}\left|\widehat{Y}_{bhi}-Y_{bhi}\right|}
	{\sum_{b,h\in\mathcal{H},i}M_{bhi}+\epsilon},
	\label{eq:pred_loss}
\end{equation}
where $\epsilon$ is a small constant to avoid division by zero. This design ensures that invalid records do not contribute to optimization and that the training objective remains consistent with the reported forecasting horizons. The backbone also returns an auxiliary frequency alignment loss inherited from the periodic-residual framework~\cite{shao2026hyperd}. Specifically, the prediction is decomposed by FFT into low-frequency and high-frequency components, and the periodic output and residual output are aligned with these components by mean squared error. Let $\mathcal{L}_{low}$ and $\mathcal{L}_{high}$ denote these two inherited losses. The training objective is
\begin{equation}
\mathcal{L}
=\mathcal{L}_{pred}
+\alpha\left(\mathcal{L}_{low}+\mathcal{L}_{high}\right),
\label{eq:total_loss}
\end{equation}
where the final configuration uses $\alpha=2.0$. This auxiliary loss is part of the inherited backbone and is not introduced as a contribution of this paper.

\section{Experiments}

\subsection{Dataset and Experimental Settings}

The experiments are conducted on the Beijing PM$_{2.5}$ dataset used in DeepPM$_{2.5}$~\cite{deeppm25}. The dataset contains air-quality and meteorological observations from 39 monitoring stations in and around Beijing, China. According to the DeepPM$_{2.5}$ dataset description, the observations cover the period from January 1, 2018 00:00 to December 31, 2020 00:00 with an hourly resolution. The 12 physical variables are PM$_{2.5}$, PM$_{10}$, SO$_2$, NO$_2$, CO, O$_3$, AQI, surface pressure, air temperature, dew-point temperature, wind direction, and average wind speed. In the current codebase, these variables correspond to the stored fields AQI, CO, NO$_2$, O$_3$, PM$_{10}$, SO$_2$, dew-point temperature, ground pressure, temperature, wind direction, wind speed, and PM$_{2.5}$; the code additionally stores temporal fields including hour, day of week, and month. The implemented model uses the 12 value features together with hour-of-day and day-of-week indicators, and predicts PM$_{2.5}$ only.

The train, validation, and test subsets follow a 60\%/20\%/20\% chronological split. The current processed tensor recorded in the project configuration contains 25,593 hourly time steps, 39 stations, and 15 raw fields. The input sequence length is 48 hours and the prediction horizon is 48 hours. The current implementation uses channel-wise z-score normalization with rescaling during evaluation. The main evaluation horizons are 1, 3, 6, 12, 24, and 48 hours. Table~\ref{tab:dataset_summary} summarizes the dataset, variables, split, prediction setting, and evaluation metrics used in the experiments.

\begin{table}[!t]
\centering
\caption{Dataset and Experimental Setting Summary}
\label{tab:dataset_summary}
\footnotesize
\resizebox{\columnwidth}{!}{%
\begin{tabular}{ll}
\hline
Item & Setting \\
\hline
Dataset & Beijing PM$_{2.5}$ dataset \\
Stations & 39 monitoring stations \\
Time range & Jan. 1, 2018 00:00 to Dec. 31, 2020 00:00 \\
Frequency & Hourly \\
Physical variables &
\begin{tabular}[t]{@{}l@{}}
	PM$_{2.5}$, PM$_{10}$, SO$_2$, NO$_2$, CO, O$_3$, AQI \\
	pressure, temperature, dew point, wind direction, wind speed
\end{tabular} \\
Code temporal fields & hour, day of week, month \\
Split & 60\% training, 20\% validation, 20\% testing \\
Input length / horizon & 48 hours / 48 hours \\
Metrics & MAE, RMSE\\
\hline
\end{tabular}%
}
\end{table}

\subsection{Baselines}

The main comparison uses the baseline results publicly reported in DeepPM$_{2.5}$~\cite{deeppm25} under the same Beijing dataset and experimental protocol. These cited rows include temporal sequence models, transformer-based models, spatio-temporal graph models, and the DeepPM$_{2.5}$ model: LSTM~\cite{hochreiter1997lstm}, GRU~\cite{cho2014gru}, Informer~\cite{zhou2021informer}, iTransformer~\cite{liu2024itransformer}, Graph-WaveNet~\cite{wu2019graphwavenet}, MTGNN~\cite{wu2020mtgnn}, DSTGN~\cite{diao2019dstgcn}, MSTGAN~\cite{zhou2024mstgan}, and DeepPM$_{2.5}$~\cite{deeppm25}. We additionally include HyperD~\cite{shao2026hyperd} as the inherited periodic-residual backbone reproduced in our codebase. Since the reported DeepPM$_{2.5}$ table provides MAE and RMSE only, the main comparison is restricted to these two metrics. 

\subsection{Comparisons with Deep Learning Methods}

The main comparison is presented from two perspectives. We first use representative prediction cases to examine whether the model follows realistic temporal changes at individual stations. We then report average and horizon-wise quantitative results to compare MVG-KAN with representative baselines.

\subsubsection{Qualitative Analysis}

\begin{figure}[!htbp]
\centering
\includegraphics[width=\columnwidth]{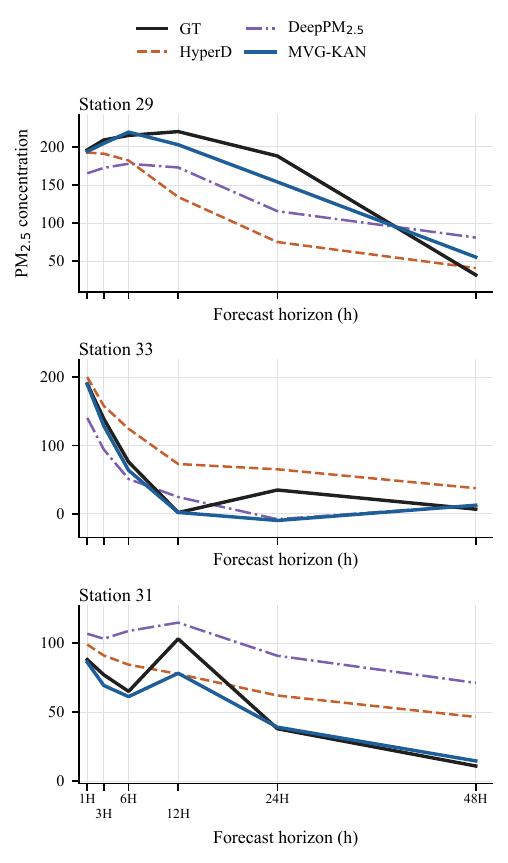}
\caption{Selected PM$_{2.5}$ forecasting examples at three monitoring stations and six evaluation horizons. The curves compare the ground truth (GT), HyperD, DeepPM$_{2.5}$, and MVG-KAN under aligned test samples.}
\label{fig:prediction_examples}
\end{figure}

Fig.~\ref{fig:prediction_examples} shows three station-level examples selected from the aligned test predictions. At Station 29, the ground truth remains high in the early horizons and then decreases toward the 48H horizon. HyperD tends to dissipate the concentration too rapidly, while DeepPM$_{2.5}$ preserves a higher long-horizon level but underestimates the early high-concentration segment. MVG-KAN follows the early peak and subsequent decline more closely. This behavior is consistent with the role of TKAN, which uses de-periodized PM$_{2.5}$ residuals and multi-pollutant histories to model local residual persistence and gradual dissipation. At Station 33, the concentration drops sharply after the first few horizons. MVG-KAN captures this rapid local transition better than HyperD, whose prediction remains too high after the drop. This suggests that modeling the residual after removing daily and weekly periodicity helps the model respond to non-periodic local changes rather than relying mainly on inherited periodic patterns. At Station 31, MVG-KAN follows the later decreasing trend, but it underestimates the abrupt 12H spike. This case is useful because it shows both the benefit and the limitation of the proposed design: TKAN improves station-wise residual correction and the Geo-Wind Graph provides transport-aware spatial context, but extremely abrupt local peaks may still be difficult when they are weakly supported by recent multi-pollutant and wind-guided signals.

\subsubsection{Quantitative Analysis}

\begin{figure}[!htbp]
	\centering
	\includegraphics[width=\columnwidth]{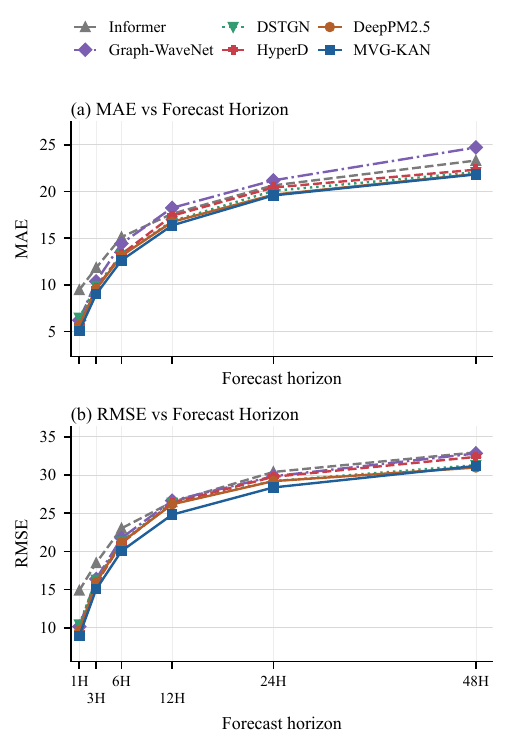}
	\caption{Horizon-wise forecasting performance comparison among representative methods on the Beijing PM$_{2.5}$ dataset. The selected methods include Informer, Graph-WaveNet, DSTGN, HyperD, DeepPM$_{2.5}$, and the proposed MVG-KAN. MVG-KAN achieves lower errors across most forecasting horizons.}
	\label{fig:horizon_performance}
\end{figure}

Table~\ref{tab:main_comparison} reports the average MAE/RMSE comparison. Rows except HyperD and MVG-KAN are cited from DeepPM$_{2.5}$~\cite{deeppm25}; HyperD and MVG-KAN use saved test metrics from the current project. Lower values indicate better forecasting performance. The horizon-wise behavior of representative methods, including DeepPM$_{2.5}$ and MVG-KAN, is further visualized in Fig.~\ref{fig:horizon_performance}.

\begin{table}[!htbp]
\centering
\caption{Performance comparison on the Beijing PM$_{2.5}$ dataset. Rows except HyperD and MVG-KAN are cited from DeepPM$_{2.5}$ under the same dataset and experimental protocol; HyperD and MVG-KAN are evaluated in our codebase.}
\label{tab:main_comparison}
\begin{tabular}{lcc}
\hline
Models & Avg MAE & Avg RMSE \\
\hline
LSTM & 17.41 & 26.29 \\
GRU & 17.25 & 26.35 \\
Informer & 13.34 & 24.37 \\
iTransformer & 16.72 & 24.27 \\
Graph-WaveNet & 15.87 & 22.93 \\
MTGNN & 16.58 & 23.23 \\
DSTGN & 14.76 & 22.47 \\
MSTGAN & 15.93 & 23.11 \\
HyperD & 14.79 & 22.46 \\
DeepPM$_{2.5}$ & 14.45 & 22.18 \\
MVG-KAN & \textbf{14.09} & \textbf{21.40} \\
\hline
\end{tabular}
\end{table}

Table~\ref{tab:main_comparison} shows that MVG-KAN achieves the lowest average errors in the comparison, with an Avg MAE of 14.09 and an Avg RMSE of 21.40. Compared with DeepPM$_{2.5}$, the Avg MAE is reduced from 14.45 to 14.09, and the Avg RMSE is reduced from 22.18 to 21.41. Compared with the inherited HyperD backbone, MVG-KAN also reduces Avg MAE from 14.79 to 14.09 and Avg RMSE from 22.46 to 21.41, indicating that the proposed PM$_{2.5}$-oriented modules improve the underlying periodic-residual foundation. Fig.~\ref{fig:horizon_performance} further compares MVG-KAN with representative Transformer-, graph-, inherited HyperD-, and PM$_{2.5}$-oriented baselines. The improvement over DeepPM$_{2.5}$ is most evident at short and medium forecasting horizons from 1H to 12H, where MVG-KAN consistently yields lower MAE and RMSE. At longer horizons, the two methods become closer, and DeepPM$_{2.5}$ remains slightly lower for 48H RMSE. Overall, these results support the complementary benefits of the transport-aware Geo-Wind Graph and the TKAN residual temporal head: the former provides physically motivated spatial guidance for residual propagation, while the latter strengthens station-wise local residual dynamics modeling through nonlinear multi-pollutant temporal correction.

\subsection{Ablation Study}

Table~\ref{tab:ablation} reports the controlled ablation results from the current project. The baseline row uses PM$_{2.5}$ as the only value input and disables the Geo-Wind Graph, TKAN, and residual AR setting. The other rows use multi-pollutant and meteorological inputs with residual AR enabled, and isolate the effects of the Geo-Wind Graph and TKAN. The ablation results are our own test results.

\begin{table}[H]
\centering
\caption{Ablation Study of the Proposed Components}
\label{tab:ablation}
\scriptsize
\resizebox{\columnwidth}{!}{%
\begin{tabular}{lcccc}
\hline
Variant  & Graph & TKAN  & MAE & RMSE \\
\hline
Baseline  & Off & Off  & 14.79 & 22.45  \\
+ Geo-Wind Graph & On & Off  & 14.69 & 22.21  \\
+ TKAN  & Off & On  & 14.28 & 21.70  \\
MVG-KAN  & On & On  & \textbf{14.09} & \textbf{21.40} \\
\hline
\end{tabular}%
}
\end{table}

Adding only the Geo-Wind Graph reduces MAE from 14.79 to 14.69 and RMSE from 22.45 to 22.21, indicating that geography- and wind-guided spatial propagation provides a useful but moderate enhancement. Adding TKAN without the Geo-Wind Graph reduces MAE to 14.28 and RMSE to 21.70, showing that station-wise KAN residual autoregression and multi-pollutant co-variation provide a stronger local temporal correction path. The full MVG-KAN model further improves MAE to 14.09 and RMSE to 21.4062. This result supports the complementary roles of the two proposed components: the Geo-Wind Graph supplies transport-aware inter-station residual propagation, while TKAN strengthens intra-station residual dynamics after the periodic component has been removed.

\subsection{Station-Wise and Distributional Diagnostics}

\begin{figure}[!htbp]
	\centering
	\includegraphics[width=\columnwidth]{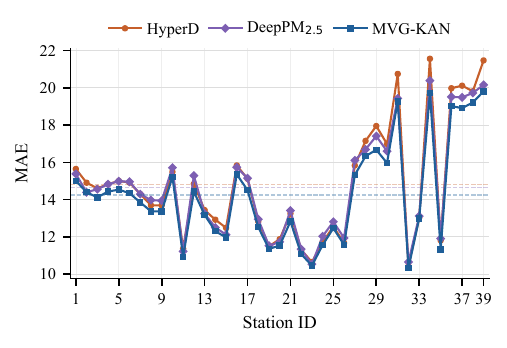}
	\caption{Station-wise MAE comparison computed from aligned test predictions at the selected evaluation horizons. The figure compares HyperD, DeepPM$_{2.5}$, and MVG-KAN across the 39 monitoring stations.}
	\label{fig:stationwise_mae}
\end{figure}

Fig.~\ref{fig:stationwise_mae} compares station-wise MAE across the 39 monitoring stations. MVG-KAN obtains lower errors than HyperD and DeepPM$_{2.5}$ at most stations, while all methods show larger errors at several high-variance stations. This station-wise pattern is consistent with the main comparison: replacing raw local temporal prediction with de-periodized TKAN residual correction improves most stations, but the remaining hard stations indicate that some local pollution episodes are still difficult to capture. The Geo-Wind Graph is expected to help when station interactions are supported by geographic proximity and wind direction, but its effect can vary spatially because the Beijing monitoring network only sparsely samples possible transport pathways.

Two observations are particularly relevant to the proposed decomposition view. First, the reduction of station-wise error is not concentrated at a single station. MVG-KAN tends to track the lower envelope of the compared methods over many low- and medium-error stations, suggesting that TKAN provides a broadly useful local residual correction after daily and weekly periodicity has been removed. This agrees with the ablation result, where TKAN contributes the larger numerical improvement. Second, the stations with the largest errors are often difficult for all methods simultaneously. Such cases are likely associated with abrupt local events, sparse upstream evidence, or pollutant changes that are weakly represented by recent multi-pollutant and wind variables. In these cases, the Geo-Wind Graph can provide a transport-aware spatial prior, but it cannot fully compensate for unobserved emissions or unresolved meteorological processes. Therefore, Fig.~\ref{fig:stationwise_mae} supports a balanced interpretation: the proposed modules improve the robustness of station-level forecasting across the network, while the remaining station-wise peaks identify where additional meteorological, emission, or remote-sensing information may be needed.

\begin{figure}[!htbp]
\centering
\includegraphics[width=\columnwidth]{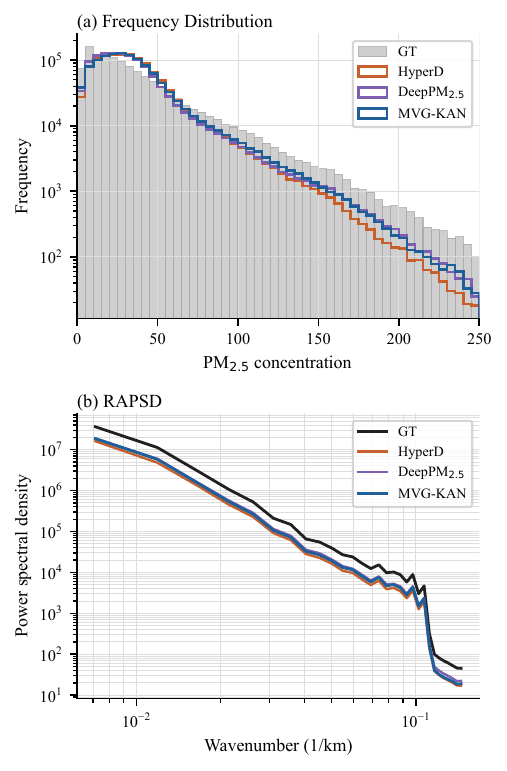}
\caption{Distributional and spatial-spectrum diagnostics of PM$_{2.5}$ forecasts. (a) Frequency distributions compare GT and model-predicted PM$_{2.5}$ concentrations at the selected evaluation horizons. (b) RAPSD compares the radially averaged spatial power spectral density after interpolating station observations onto a regular geographic grid.}
\label{fig:distribution_rapsd}
\end{figure}

Fig.~\ref{fig:distribution_rapsd} provides two additional descriptive diagnostics. In the frequency distribution, MVG-KAN follows the overall PM$_{2.5}$ concentration distribution of the ground truth and better preserves the middle-to-high concentration range than the compared prediction files. This is relevant to TKAN because residual correction is designed to retain short-term pollutant accumulation and slow dissipation after periodicity is removed. In the RAPSD plot, MVG-KAN stays close to the ground-truth curve over much of the low- and middle-wavenumber range, suggesting that the broad spatial variability of the station field is reasonably preserved after interpolation. This observation is consistent with the role of the Geo-Wind Graph as a lightweight spatial prior. However, RAPSD here should be interpreted only as a diagnostic visualization because it is computed after interpolating irregular station observations onto a regular grid, not as direct evidence that the model physically simulates atmospheric transport.

\section{Discussion}

The experimental results support the proposed forecasting view that station-level PM$_{2.5}$ should be modeled through local periodic patterns, local residual dynamics, and wind-driven spatial transport. The qualitative examples in Fig.~\ref{fig:prediction_examples} show that MVG-KAN can better follow several local concentration transitions after periodic effects are removed. The quantitative comparison in Table~\ref{tab:main_comparison} and Fig.~\ref{fig:horizon_performance} shows lower average MAE and RMSE than the cited baselines across most forecasting horizons, and the ablation study in Table~\ref{tab:ablation} further separates the roles of the two proposed modules. TKAN provides the stronger local residual temporal correction, while the Geo-Wind Graph contributes a complementary transport-aware spatial prior. The station-wise, distributional, and RAPSD diagnostics in Figs.~\ref{fig:stationwise_mae} and~\ref{fig:distribution_rapsd} also suggest that the improvements are not limited to a single averaged metric, although the gains vary across stations and horizons.

Several limitations remain. First, the Geo-Wind Graph is a lightweight physically motivated graph rather than a full atmospheric transport simulator. It depends on the quality and spatial representativeness of the available wind variables, and the 39-station network may not fully capture transport pathways through unmonitored areas. Second, the current graph does not explicitly model emissions, boundary-layer height, vertical mixing, deposition, or chemical formation, which may be important during severe or chemically complex pollution episodes. Third, TKAN is intentionally station-wise; it improves local residual correction but does not itself model spatial propagation. Finally, the main comparison uses several baseline rows reported by DeepPM$_{2.5}$ rather than fully reproduced under one codebase, so the comparison should be interpreted with this practical constraint.

Future work will investigate more refined meteorological fields, trajectory-based transport features, boundary-layer information, emission inventories, remote sensing observations, and chemical-transport-model outputs. More extensive reproduced baselines, repeated runs, uncertainty analysis, and cross-city or cross-season evaluation would also help clarify the robustness of MVG-KAN across different pollution regimes and forecasting horizons.

\section{Conclusion}

Short-term PM$_{2.5}$ forecasting is challenging because observed concentrations are affected by temporal periodicity, non-periodic pollution episodes, wind-driven transport, and multi-pollutant co-variation. This paper presented a geo-wind guided KAN residual temporal learning framework for station-level PM$_{2.5}$ forecasting. The framework uses a periodic-residual prediction backbone as its foundation and introduces two PM$_{2.5}$-oriented components: a Geo-Wind Graph and a TKAN residual temporal module.

The Geo-Wind Graph combines geographic proximity and wind-driven directionality to enhance physically meaningful spatial dependency modeling among monitoring stations. It uses station locations and wind variables to guide directed residual information flow, so that spatial interaction is more consistent with potential pollutant transport. TKAN uses the de-periodized PM$_{2.5}$ residual and historical multi-pollutant sequences for station-wise nonlinear residual correction. It does not perform spatial propagation; instead, it complements the graph module by modeling local residual inertia, abrupt temporal changes, and pollutant co-variation.

Experiments on the Beijing PM$_{2.5}$ dataset show that MVG-KAN achieves MAE 14.09 and RMSE 21.40. The ablation study indicates that TKAN provides the major performance gain, reducing MAE and RMSE more clearly than the graph-only variant. The Geo-Wind Graph provides complementary spatial enhancement, with moderate improvements in MAE and RMSE in the graph-only setting. These results suggest that station-wise KAN residual temporal correction and physically guided spatial propagation play different but compatible roles in PM$_{2.5}$ forecasting.

Overall, the results indicate that separating local residual temporal correction from wind-guided spatial propagation is a practical direction for PM$_{2.5}$ forecasting.

\bibliographystyle{IEEEtran}
\bibliography{references}

\end{document}